\documentclass{article}
\usepackage{MLDD_workshop_2022,times}
\usepackage{graphicx} 
\usepackage{caption}

\usepackage{subcaption}

\usepackage{amsmath,amsfonts,bm}

\def\eqref#1{equation~\ref{#1}}

\def\1{\bm{1}}

\DeclareMathAlphabet{\mathsfit}{\encodingdefault}{\sfdefault}{m}{sl}
\SetMathAlphabet{\mathsfit}{bold}{\encodingdefault}{\sfdefault}{bx}{n}

\usepackage{amsmath}
\usepackage{amssymb}
\usepackage{amsthm}

\usepackage{hyperref}
\usepackage{url}
\usepackage{wrapfig,booktabs}
\usepackage{lipsum}

\usepackage{tikz}
\usetikzlibrary{shapes.geometric, arrows}
\tikzstyle{startstop} = [rectangle, rounded corners, text centered, draw=black, left color=green!30, right color=green!30,shading angle=0]
\tikzstyle{io} = [trapezium, trapezium left angle=70, trapezium right angle=110, minimum width=3cm, minimum height=1cm, text centered, draw=black, fill=green!30]
\tikzstyle{process} = [rectangle, minimum height=0.2cm, text centered, draw=black, left color=green!30, right color=green!30,shading angle=0]
\tikzstyle{decision} = [circle, minimum size= 0.01cm, text centered, draw=black, left color=green!30, right color=green!30,shading angle=0]
\tikzstyle{graph} = [circle, minimum size= 0.05cm, text centered, draw=black]

\tikzstyle{arrow} = [thick,-,>=stealth]

\usepackage{pgfplots}

\pgfplotsset{
        colormap={my colormap}{
            rgb255=(255, 0, 0),
            rgb255=(255, 255, 255  ),
            rgb255=(0, 255, 0),
        },
    }

\pgfplotsset{
        colormap={my colormap two}{
            rgb255=(255, 255, 255  ),
            rgb255=(0, 255, 0),
        },
    }

\title{Graph Anisotropic Diffusion}

\author{\hspace{1.5cm}Ahmed A. A. Elhag\thanks{Correspondence to \texttt{aelhag@aimsammi.org}} \\ 
\hspace{1.5cm}African Masters of Machine Intelligence\\
\And 
\hspace{-5.4cm}Gabriele Corso \\
\hspace{-5.4cm}MIT \\
\AND
\hspace{3.37cm}Hannes Stärk \\
\hspace{3.37cm}MIT \\
\And
\hspace{0.1cm}Michael M. Bronstein \\
\hspace{0.1cm}University of Oxford / Twitter\\
}
\iclrfinalcopy 
\begin{document}
\maketitle

\begin{abstract}
Traditional Graph Neural Networks (GNNs) rely on message passing, which amounts to permutation-invariant local aggregation of neighbour features. Such a process is isotropic and there is no notion of `direction' on the graph.  
We present a new GNN architecture called Graph Anisotropic Diffusion. Our model alternates between linear diffusion, for which a closed-form solution is available, and local anisotropic filters to obtain efficient multi-hop anisotropic kernels.  
%
We test our model on two common molecular property prediction benchmarks (ZINC and QM9) and show its competitive performance.
\end{abstract}

\section{Introduction}
\label{Introduction}
Graphs are a very common mathematical abstraction used to represent complex systems of relations and interactions.
However, they present several challenges for machine learning methods because they require algorithms to reason about their underlying structure as well as their node and edge features.
Graph Neural Networks (GNNs) are a popular architecture for learning on graph-structured data. GNNs have been successfully applied to problems in different domains such as  social networks, recommendation systems \citep{2018GNNs, fan2019graph}, chemistry \citep{gilmer2017neural, sanchezlengeling2019machine}, bioinformatics \citep{lv2021learning, molecule_bio}, and drug discovery \citep{gaudelet2021utilising}. 

The majority of GNNs are based on the {\em message passing} mechanism \citep{gilmer2017neural}, in which the features of each node in the graph are updated by aggregating information from its neighbours. 
However, recent studies have demonstrated several limitations of message passing, which include limited expressive power (due to equivalence to the Weisfeiler-Lehman graph isomorphism test) \citep{xu2019powerful, corso2020principal, bodnar2021weisfeiler}, as well as the oversmoothing \citep{oono2021graph} and bottleneck phenomena \citep{alon2021bottleneck}.

A recent line of works linked GNNs to diffusion processes on graphs. GRAND \citep{chamberlain2021grand} and BLEND \citep{chamberlain2021beltrami} obtained GNNs from a discretisation of nonlinear and non-Euclidean diffusion PDEs. 
A more general version of diffusion-based on cellular sheaves constructed upon graphs was studied by  \cite{bodnar2022neural}.
%
\cite{klicpera2019diffusion} applied diffusion to the graph structure itself, as a form of graph rewiring that alleviates bottlenecks and improved learning in homophilic settings. 
\cite{topping2021understanding} related bottlenecks to graph curvature and used a discrete version of Ricci flow (which, in the continuous setting, can be conceptually interpreted as the `diffusion of the metric') to improve oversquashing in GNNs.




The lack of canonical ordering of graph nodes forces message passing to use permutation-invariant functions, meaning there is no natural `direction' on graphs. 
In geometric graphs and meshes where additional structure is available, learning schemes based on anisotropic diffusion have been studied \citep{andreux2014anisotropic,boscaini2016anisotropic,boscaini2016learning,sharp2021diffusionnet}. %
%
%
Attempts to define `pseudodirections' in GNNs using graph motifs were undertaken by \cite{monti2018motifnet}. More recently, 
\citep{beaini2021directional} proposed a scheme for adding directionality in message passing using the graph Laplacian eigenvectors. 
%

\begin{figure}[t]
\centering
\hspace{-1.7cm}
\subfloat[Diffusion for different times $t$]{
\begin{tikzpicture}[node distance=0cm]

 
\node (graph_node_1) [graph, fill=green!56.9] {};
\node (graph_node_2) [graph, shift=({-30:0.5cm}), fill=green!15.33] {};
\node (graph_node_3) [graph, right of=graph_node_2, shift=({30:0.5 cm}), fill=green!4.38] {};
\node (graph_node_4) [graph, right of=graph_node_3, shift=({90:0.5cm}), fill=green!2.19] {};
\node (graph_node_5) [graph, right of=graph_node_4, shift=({150:0.5cm}), fill=green!4.38] {};
\node (graph_node_6) [graph, shift=({90:0.5cm}), fill=green!15.32] {};
\draw [thick,-,>=stealth] (graph_node_1) -- (graph_node_2);
\draw [thick,-,>=stealth] (graph_node_1) -- (graph_node_6);
\draw [thick,-,>=stealth] (graph_node_2) -- (graph_node_3);
\draw [thick,-,>=stealth] (graph_node_6) -- (graph_node_5);
\draw [thick,-,>=stealth] (graph_node_5) -- (graph_node_4);
\draw [thick,-,>=stealth] (graph_node_3) -- (graph_node_4);

\node (graph_node_7) [graph, shift=({210:0.5cm}), fill=green!10.88] {};
\node (graph_node_8) [graph, right of=graph_node_7, shift=({-90:0.5cm}), fill=green!5.44] {};
\node (graph_node_9) [graph, right of=graph_node_7, shift=({150:0.5cm}), fill=green!2.91] {};

\draw [thick,-,>=stealth] (graph_node_1) -- (graph_node_7);
\draw [thick,-,>=stealth] (graph_node_7) -- (graph_node_8);
\draw [thick,-,>=stealth] (graph_node_7) -- (graph_node_9);

\node (graph_node_10) [graph, right of=graph_node_9, shift=({210:0.5cm}), fill=green!0.77] {};
\draw [thick,-,>=stealth] (graph_node_9) -- (graph_node_10);

\node (graph_node_11) [graph, right of=graph_node_10, shift=({150:0.5cm}), fill=green!0.15] {};
\draw [thick,-,>=stealth] (graph_node_10) -- (graph_node_11);

\node (graph_node_12) [graph, right of=graph_node_11, shift=({90:0.5cm}), fill=green!0.07] {};
\node (graph_node_13) [graph, right of=graph_node_11, shift=({210:0.5cm}), fill=green!0.07] {};

\draw [thick,-,>=stealth] (graph_node_11) -- (graph_node_12);
\draw [thick,-,>=stealth] (graph_node_11) -- (graph_node_13);

\node (value) [right of=graph_node_6, shift=({180:0.75cm})] {$t = 1$};

\end{tikzpicture}
\hspace{0.2cm}
\begin{tikzpicture}[node distance=0cm]



\node (graph_node_1) [graph, fill=green!29.75] {};
\node (graph_node_2) [graph, shift=({-30:0.5cm}), fill=green!16.88] {};
\node (graph_node_3) [graph, right of=graph_node_2, shift=({30:0.5 cm}), fill=green!10.78] {};
\node (graph_node_4) [graph, right of=graph_node_3, shift=({90:0.5cm}), fill=green!8.98] {};
\node (graph_node_5) [graph, right of=graph_node_4, shift=({150:0.5cm}), fill=green!10.78] {};
\node (graph_node_6) [graph, shift=({90:0.5cm}), fill=green!16.88] {};
\draw [thick,-,>=stealth] (graph_node_1) -- (graph_node_2);
\draw [thick,-,>=stealth] (graph_node_1) -- (graph_node_6);
\draw [thick,-,>=stealth] (graph_node_2) -- (graph_node_3);
\draw [thick,-,>=stealth] (graph_node_6) -- (graph_node_5);
\draw [thick,-,>=stealth] (graph_node_5) -- (graph_node_4);
\draw [thick,-,>=stealth] (graph_node_3) -- (graph_node_4);

\node (graph_node_7) [graph, shift=({210:0.5cm}), fill=green!13.33] {};
\node (graph_node_8) [graph, right of=graph_node_7, shift=({-90:0.5cm}), fill=green!11.11] {};
\node (graph_node_9) [graph, right of=graph_node_7, shift=({150:0.5cm}), fill=green!7.13] {};

\draw [thick,-,>=stealth] (graph_node_1) -- (graph_node_7);
\draw [thick,-,>=stealth] (graph_node_7) -- (graph_node_8);
\draw [thick,-,>=stealth] (graph_node_7) -- (graph_node_9);

\node (graph_node_10) [graph, right of=graph_node_9, shift=({210:0.5cm}), fill=green!3.78] {};
\draw [thick,-,>=stealth] (graph_node_9) -- (graph_node_10);

\node (graph_node_11) [graph, right of=graph_node_10, shift=({150:0.5cm}), fill=green!1.95] {};
\draw [thick,-,>=stealth] (graph_node_10) -- (graph_node_11);

\node (graph_node_12) [graph, right of=graph_node_11, shift=({90:0.5cm}), fill=green!1.63] {};
\node (graph_node_13) [graph, right of=graph_node_11, shift=({210:0.5cm}), fill=green!1.63] {};

\draw [thick,-,>=stealth] (graph_node_11) -- (graph_node_12);
\draw [thick,-,>=stealth] (graph_node_11) -- (graph_node_13);

\node (value) [right of=graph_node_6, shift=({180:0.75cm})] {$t = 5$};

\end{tikzpicture}
\newline
\begin{tikzpicture}[node distance=0cm]

\node (graph_node_1) [graph, fill=green!100] {};
\node (graph_node_2) [graph, shift=({-30:0.5cm})] {};
\node (graph_node_3) [graph, right of=graph_node_2, shift=({30:0.5 cm})] {};
\node (graph_node_4) [graph, right of=graph_node_3, shift=({90:0.5cm})] {};
\node (graph_node_5) [graph, right of=graph_node_4, shift=({150:0.5cm})] {};
\node (graph_node_6) [graph, shift=({90:0.5cm}) fill=red!60] {};
\draw [thick,-,>=stealth] (graph_node_1) -- (graph_node_2);
\draw [thick,-,>=stealth] (graph_node_1) -- (graph_node_6);
\draw [thick,-,>=stealth] (graph_node_2) -- (graph_node_3);
\draw [thick,-,>=stealth] (graph_node_6) -- (graph_node_5);
\draw [thick,-,>=stealth] (graph_node_5) -- (graph_node_4);
\draw [thick,-,>=stealth] (graph_node_3) -- (graph_node_4);

\node (graph_node_7) [graph, shift=({210:0.5cm})] {};
\node (graph_node_8) [graph, right of=graph_node_7, shift=({-90:0.5cm})] {};
\node (graph_node_9) [graph, right of=graph_node_7, shift=({150:0.5cm})] {};

\draw [thick,-,>=stealth] (graph_node_1) -- (graph_node_7);
\draw [thick,-,>=stealth] (graph_node_7) -- (graph_node_8);
\draw [thick,-,>=stealth] (graph_node_7) -- (graph_node_9);

\node (graph_node_10) [graph, right of=graph_node_9, shift=({210:0.5cm})] {};
\draw [thick,-,>=stealth] (graph_node_9) -- (graph_node_10);

\node (graph_node_11) [graph, right of=graph_node_10, shift=({150:0.5cm})] {};
\draw [thick,-,>=stealth] (graph_node_10) -- (graph_node_11);

\node (graph_node_12) [graph, right of=graph_node_11, shift=({90:0.5cm})] {};
\node (graph_node_13) [graph, right of=graph_node_11, shift=({210:0.5cm})] {};

\draw [thick,-,>=stealth] (graph_node_11) -- (graph_node_12);
\draw [thick,-,>=stealth] (graph_node_11) -- (graph_node_13);

\node (value) [right of=graph_node_6, shift=({180:0.75cm})] {$t = 0$};

\end{tikzpicture}

\label{fig:diffusion kernel diagram}}
\hspace{-3cm}
\subfloat[Anisotropic filters: $av_{1}$ and $dx_{1}$]{
\hspace{0.6cm}
\begin{tikzpicture}[node distance=0cm]

\node (graph_node_1) [graph, fill=black!100] {};
\node (graph_node_2) [graph, shift=({-30:0.5cm}), fill=green!63] {};
\node (graph_node_3) [graph, right of=graph_node_2, shift=({30:0.5 cm})] {};
\node (graph_node_4) [graph, right of=graph_node_3, shift=({90:0.5cm})] {};
\node (graph_node_5) [graph, right of=graph_node_4, shift=({150:0.5cm})] {};
\node (graph_node_6) [graph, shift=({90:0.5cm}), fill=green!63] {};
\draw [thick,-,>=stealth] (graph_node_1) -- (graph_node_2);
\draw [thick,-,>=stealth] (graph_node_1) -- (graph_node_6);
\draw [thick,-,>=stealth] (graph_node_2) -- (graph_node_3);
\draw [thick,-,>=stealth] (graph_node_6) -- (graph_node_5);
\draw [thick,-,>=stealth] (graph_node_5) -- (graph_node_4);
\draw [thick,-,>=stealth] (graph_node_3) -- (graph_node_4);

\node (graph_node_7) [graph, shift=({210:0.5cm}), fill=green!50] {};
\node (graph_node_8) [graph, right of=graph_node_7, shift=({-90:0.5cm})] {};
\node (graph_node_9) [graph, right of=graph_node_7, shift=({150:0.5cm})] {};

\draw [thick,-,>=stealth] (graph_node_1) -- (graph_node_7);
\draw [thick,-,>=stealth] (graph_node_7) -- (graph_node_8);
\draw [thick,-,>=stealth] (graph_node_7) -- (graph_node_9);

\node (graph_node_10) [graph, right of=graph_node_9, shift=({210:0.5cm})] {};
\draw [thick,-,>=stealth] (graph_node_9) -- (graph_node_10);

\node (graph_node_11) [graph, right of=graph_node_10, shift=({150:0.5cm})] {};
\draw [thick,-,>=stealth] (graph_node_10) -- (graph_node_11);

\node (graph_node_12) [graph, right of=graph_node_11, shift=({90:0.5cm})] {};
\node (graph_node_13) [graph, right of=graph_node_11, shift=({210:0.5cm})] {};

\draw [thick,-,>=stealth] (graph_node_11) -- (graph_node_12);
\draw [thick,-,>=stealth] (graph_node_11) -- (graph_node_13);

\node (value) [right of=graph_node_6, shift=({180:0.75cm})] {$av_{1}$};

\end{tikzpicture}
\newline
\hspace{-1.5cm}
\begin{tikzpicture}[node distance=0cm]

\node (graph_node_1) [graph, fill=black!100] {};
\node (graph_node_2) [graph, shift=({-30:0.5cm}), fill=red!63] {};
\node (graph_node_3) [graph, right of=graph_node_2, shift=({30:0.5 cm})] {};
\node (graph_node_4) [graph, right of=graph_node_3, shift=({90:0.5cm})] {};
\node (graph_node_5) [graph, right of=graph_node_4, shift=({150:0.5cm})] {};
\node (graph_node_6) [graph, shift=({90:0.5cm}), fill=red!63] {};
\draw [thick,-,>=stealth] (graph_node_1) -- (graph_node_2);
\draw [thick,-,>=stealth] (graph_node_1) -- (graph_node_6);
\draw [thick,-,>=stealth] (graph_node_2) -- (graph_node_3);
\draw [thick,-,>=stealth] (graph_node_6) -- (graph_node_5);
\draw [thick,-,>=stealth] (graph_node_5) -- (graph_node_4);
\draw [thick,-,>=stealth] (graph_node_3) -- (graph_node_4);

\node (graph_node_7) [graph, shift=({210:0.5cm}), fill=green!50] {};
\node (graph_node_8) [graph, right of=graph_node_7, shift=({-90:0.5cm})] {};
\node (graph_node_9) [graph, right of=graph_node_7, shift=({150:0.5cm})] {};

\draw [thick,-,>=stealth] (graph_node_1) -- (graph_node_7);
\draw [thick,-,>=stealth] (graph_node_7) -- (graph_node_8);
\draw [thick,-,>=stealth] (graph_node_7) -- (graph_node_9);

\node (graph_node_10) [graph, right of=graph_node_9, shift=({210:0.5cm})] {};
\draw [thick,-,>=stealth] (graph_node_9) -- (graph_node_10);

\node (graph_node_11) [graph, right of=graph_node_10, shift=({150:0.5cm})] {};
\draw [thick,-,>=stealth] (graph_node_10) -- (graph_node_11);

\node (graph_node_12) [graph, right of=graph_node_11, shift=({90:0.5cm})] {};
\node (graph_node_13) [graph, right of=graph_node_11, shift=({210:0.5cm})] {};

\draw [thick,-,>=stealth] (graph_node_11) -- (graph_node_12);
\draw [thick,-,>=stealth] (graph_node_11) -- (graph_node_13);

\node (value) [right of=graph_node_6, shift=({180:0.75cm})] {$dx_{1}$};

\end{tikzpicture}

 \label{fig:anisotropic filters diagram}}
\hspace{-1cm}
\subfloat{
\centering
\begin{tikzpicture}
\pgfplotscolorbardrawstandalone[ 
    colormap name = my colormap,
    colorbar horizontal,
    point meta min=-1,
    point meta max=1,
    colorbar style={
        width=3.3cm,
        height=0.2cm,
        rotate=90,
        x tick label style={font=\tiny, xshift = 0.1cm, yshift = 0.2cm, text width=0.05cm,align=right},
        xtick={0},
        }]
\node (max) [xshift= 0.85cm, yshift = -0.1cm] {\small{max}};
\node (min) [xshift= 0.85cm, yshift = -3.18cm] {\small{min}};
\end{tikzpicture}
}
\caption{Diffusion kernels and anisotropic filters on a molecular graph. (a) Diffusion for different times $t$ of an initial scalar feature localized at node shown in green at time $t = 0$. (b) Local directional smoothing and directional derivative for the black node.} 
\label{fig:filters}
\end{figure}
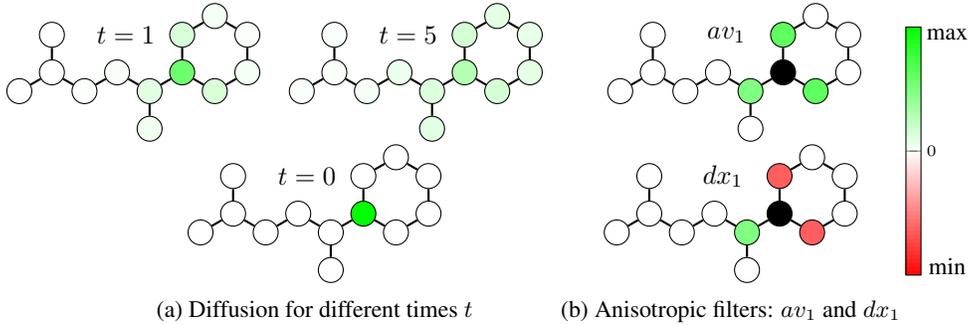

In this work, we build on the recent works of \cite{sharp2021diffusionnet} and \cite{beaini2021directional} to develop anisotropic diffusion on graphs as a basis for GNNs. We propose a new way to perform isotropic diffusion on graphs with learnable kernel size (see Figure \ref{fig:diffusion kernel diagram}) that can be applied efficiently by either solving a linear system or through spectral expansion. We combine this diffusion with a local anisotropic filter (see Figure \ref{fig:anisotropic filters diagram}) to obtain a global anisotropic kernel over the graph. This technique is leveraged to propose a novel GNN architecture that we call {\em Graph Anisotropic Diffusion} (GAD). 

Empirically, we show that GAD improves the performance of competitive GNNs on two popular molecular property prediction benchmarks: ZINC and QM9. 
\section{Background}
\label{Background}
\textbf{Diffusion equation.}
\label{Diffusion equation}
Diffusion describes the movement of some quantity from regions of high concentration to lower concentration over time. A classical example of this phenomenon is the diffusion of heat on an iron rod -- heat diffuses from warmer parts of the rod to colder ones.  
%
Given a domain $\Omega$ (e.g., a 1D iron rod), let $x(u,t) : \Omega\times [0,\infty) \rightarrow \mathbb{R}$ be a function representing some quantity (e.g., temperature) at position $u \in \Omega$ and time $t \geq 0$.
%
%
The diffusion process is governed by a partial differential equation (PDE) called the {\em heat equation}, 
\begin{equation}\label{diffusion PDE}
    \frac{\partial}{\partial t}x(u, t) = \frac{\partial^2}{\partial u^2}x(u, t) = \Delta x(u, t),
\end{equation}
where $\Delta$ is the \textit{Laplacian operator} associated with the domain.\footnote{On domains with non-Euclidean geometry, the Laplacian is referred to as the Laplace-Beltrami operator. }  
Assuming initial condition $x(u, 0)$, the solution of Equation \ref{diffusion PDE} is given by the heat operator $x(u, t) = \exp({-t\Delta}) x(u, 0)$, where $\exp$ is the operator exponential.

\textbf{Diffusion equation on surfaces.}
\label{Diffusion equation on surfaces}
In computer graphics applications, it is common to work with discretisations of continuous domains (manifolds) such as meshes or point clouds. Let us assume that the domain is sampled at $n$ points, $\{u_1, \hdots, u_n \} \subseteq \Omega$. The sampled function can be represented as an $n$-dimensional vector $\mathbf{x} = (x(u_1), \hdots, x(u_n))$. The 
continuous Laplacian operator $\Delta$ is discretised using finite elements methods giving rise to the (weak) Laplacian matrix $\mathbf{L}$ and the mass matrix $\mathbf{M}$ (both of size $n\times n$), such that the diffusion rate is approximated by  $\mathbf{M}^{-1}\mathbf{L}\mathbf{x}$.  
Equation \ref{diffusion PDE} becomes
\begin{equation}\label{mesh PDE}
    \frac{\partial}{\partial t}\mathbf{x}(t) = \mathbf{M}^{-1}\mathbf{L}\mathbf{x}(t),
\end{equation}
where $\mathbf{x}(t)$ represents the temperature on the nodes of the discretised domain at time $t$.
On meshes, it is customary to use the \textit{cotangent scheme} for Laplacian discretisation and lumped (diagonal) mass matrix, which can be interpreted as a weighted inner product. 
%
\section{Graph Anisotropic Diffusion}
\label{Diffusion equation on graphs}
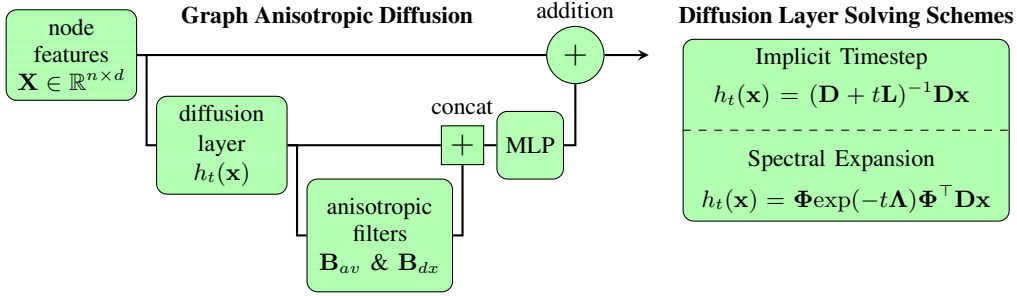
\begin{figure}[t]
\centering
\begin{tikzpicture}[node distance=1.2cm]
\node (features) [startstop, text width=1.5cm] {\small{node features\\ $\mathbf{X} \in \mathbb{R}^{n \times d}$}};
\node (diffusion) [startstop, below of=features, text width=1.5cm,  minimum height=1.2cm, xshift= 2cm] {\small{diffusion layer  $h_t(\mathbf{x})$}};

\draw [arrow] (features) -| node {} ++(1cm,0) -| node {} ++(0,-1.2cm) -| (diffusion.west) ;

\node (anisotropic) [startstop, below of=diffusion, text width=1.7cm,  minimum height=1.5cm, xshift= 2.1cm] {\small{anisotropic filters\\ $\mathbf{B}_{av}$ \& $\mathbf{B}_{dx}$}};

\draw [arrow] (diffusion) -| node {} ++(1cm,0) -| node {} ++(0,-1.2cm) -| (anisotropic.west) ;

\node (plus) [process, below of=features, xshift= 5.2cm] {\scalebox{1.2}{$+$}};
\node (plus_text) [below of=features, xshift= 5.2cm, yshift = 0.5cm] {\small{concat}};

\draw [arrow] (anisotropic) -| node{} (plus);
\draw [arrow] (diffusion) -| node{} (plus.west);

\node (mlp) [startstop, below of=features, minimum height=0.9cm, xshift= 6.1cm] {\small{MLP}};

\draw [arrow] (plus) -| node{} (mlp.west);

\node (addition) [decision,  xshift= 6.7cm] {\scalebox{1.2}{$+$}};
\node (addition_text) [xshift= 6.7cm, yshift= 0.6cm] {\small{addition}};
\node (GAD) [xshift= 3.4cm, yshift= 0.5cm] {\small{\textbf{Graph Anisotropic Diffusion}}};

\draw [arrow] (mlp) -| node{} (addition);

\draw [arrow] (features) -| node{} (addition.west);

\node (prep) [startstop , below of=features, xshift= 10.3cm, yshift= 0.2cm, text width=4.1cm] {\small{Implicit Timestep \vspace{0.15cm}\\ $h_{t}(\mathbf{x}) = (\mathbf{D} + t\mathbf{L})^{-1}\mathbf{D}\mathbf{x}$ \vspace{0.5cm} \\
Spectral Expansion \vspace{0.15cm}\\ $    h_{t}(\mathbf{x}) = \boldsymbol{\Phi}  \mathrm{exp}(-t\boldsymbol{\Lambda})
    \boldsymbol{\Phi}^\top\mathbf{D}\mathbf{x} $}};

\node (GAD_scheme) [xshift= 10.3cm, yshift= 0.5cm] {\small{\textbf{Diffusion Layer Solving Schemes }}};

\draw[dashed] (prep.west) -- (prep.east);
\node (node_1) [xshift= 7.8cm]{};
\draw [thick,->,>=stealth] (addition.east) -- (node_1);
\end{tikzpicture}
\caption{One block of Graph Anisotropic Diffusion.} 
\label{One block of Graph Anisotropic Diffusion}
\end{figure}

In this section, we present our Graph Anisotropic Diffusion (GAD) method. First, we show how a closed-form solution to the graph diffusion equation can be obtained. Then, we describe the local anisotropic kernels we use to provide our architecture with a notion of direction. Finally, we explain how these two components fit together in GAD. Figure \ref{One block of Graph Anisotropic Diffusion} provides a high-level overview of the GAD layer, and Figures \ref{fig:filters} and \ref{fig:Global Anisotropic Diffusion kernel} provide an intuition behind our construction. 
%

\subsection{Learnable Diffusion}
\label{Learned Diffusion}
Let $\mathcal{G} = (\mathcal{V}, \mathcal{E})$ be an undirected graph with $|\mathcal{V}| = n$ nodes and $|\mathcal{E}| = e$ edges, $\mathbf{A}$ be the $n \times n$ adjacency matrix of $\mathcal{G}$, and $\mathbf{D}$ is the node degree matrix. 
%
Our method is based on the discretisation of the diffusion equation (Equation \ref{diffusion PDE}) analogous to mesh diffusion equation (Equation \ref{mesh PDE}), where instead we use the graph Laplacian matrix $\mathbf{L} = \mathbf{D} - \mathbf{A}$ and the degree matrix as the mass matrix to diffuse the $n\times d$ input feature matrix $\mathbf{X}$: 
\begin{equation}\label{graph PDE}
    \frac{\partial}{\partial t}\mathbf{X}(t) = \mathbf{D}^{-1}(\mathbf{D} - \mathbf{A})\mathbf{X}(t)
\end{equation}
%
%
We use a \textit{channel-wise linear diffusion layer} $h_{t}: \mathbb{R}^{n} \rightarrow \mathbb{R}^{n}$, diffusing each channel (column of $\mathbf{X}$) by applying Equation~\ref{graph PDE} for a time $t$ that is also \textit{learnable} per channel. 
This enables us to go beyond traditional message passing in GNNs. Instead of a single propagation step or discrete random walk, we perform a continuous diffusion with control over the parameter $t$ that determines the kernel size and can be learned end-to-end. As a result, we can optimise the propagation of the features at each node from purely local to completely global by learning each channel's appropriate $t$.

\textbf{How to compute the diffusion?}
\label{How to compute the diffusion?}
Different numerical schemes are available to solve the linear diffusion equation (Equation~\ref{graph PDE}). To learn and optimise the parameter $t$ via back-propagation, we should confine ourselves to differentiable schemes.
In this regard, two simple schemes are used in the experiments: \textit{implicit timestep} and \textit{spectral expansion}. The implementation of both is simple using common scientific libraries.
Implicit timestep requires solving a large sparse linear system. In contrast, the spectral expansion scheme has a reduced time complexity and it relies on dense arithmetic but requires some precomputation. In the following, we detail both options.

\textbf{Direct implicit timestep.}
\label{Direct Implicit Timestep}
\citet{sharp2021diffusionnet} used a single implicit Euler timestep to approximate the diffusion, which in our setting amounts to  
 \begin{equation}\label{equation 3}
    h_{t}(\mathbf{x}) = (\mathbf{D} + t\mathbf{L})^{-1}\mathbf{D}\mathbf{x}
\end{equation}
This approach requires solving a sparse linear system of equations for each diffusion operation using Cholesky decomposition.

\textbf{Spectral expansion scheme.}
\label{Spectral Scheme}
Another way to approximate the solution of the diffusion equation is using spectral (Fourier) expansion. On a graph, an orthogonal Fourier basis is given by the (generalised) eigenvectors of the normalised graph Laplacian, $ \mathbf{L} \boldsymbol{\phi}_i = \lambda_i \mathbf{D} \boldsymbol{\phi}_i$. 
%
%
%
%
The solution of the diffusion equation can be expressed through the spectral expansion of the heat operator, 
%
\begin{equation}\label{equation 5}
\centering
    h_{t}(\mathbf{x}) \approx \boldsymbol{\Phi}_k \,
\mathrm{exp}(-t\boldsymbol{\Lambda}_k)
\boldsymbol{\Phi}_k^\top \mathbf{D}\mathbf{x},
\end{equation}
where $\boldsymbol{\Phi}_k = (\boldsymbol{\phi}_1,\hdots, \boldsymbol{\phi}_k)$ is the $n\times k$ matrix of the first $k$ eigenvectors and $\boldsymbol{\Lambda}_k = \mathrm{diag}(\lambda_1, \hdots, \lambda_k)$ is the $k\times k$ diagonal matrix of the corresponding eigenvalues. This is an approximation due to dropping higher-frequency components, and $k$ plays the role of bandwidth. 
%
The matrices $\boldsymbol{\Phi}_k$ and $\boldsymbol{\Lambda}_k$ can be pre-computed, as they do not depend on the features $\mathbf{x}$ nor the learnable parameter $t$. Therefore, with a small approximation error, we obtain a significantly more efficient method than the implicit Euler approach. 
\subsection{Anisotropic filters}
\label{Anisotropic filters}
\cite{beaini2021directional} proposed the use of anisotropic filters on graphs and defined them as aggregation matrices that allow directional flows between nodes in the graph. 
These matrices depend on the graph structure and are derived from a vector field $\mathbf{F}$ (represented as an $n \times n$ matrix) that associates each edge with a scalar weight describing the  `flow' from one node to its neighbours. These weights correspond to the magnitude of the flow in a specific direction.

Following \cite{beaini2021directional}, we use the gradient of the first non-constant eigenvector of the normalised graph Laplacian 
(the Fiedler vector, which we denote here by $\boldsymbol{\phi}$ )
to define the vector field 
\begin{equation}\label{vector_field}
   \mathbf{F} = \mathbf{A} \odot \nabla \boldsymbol{\phi},
\end{equation}
where $\odot$ is element-wise multiplication and $(\nabla \boldsymbol{\phi})_{ij}= \phi_{i} - \phi_{j}$ is an $n\times n$ matrix.  
\cite{beaini2021directional} employ this vector field to define two aggregation matrices $\mathbf{B}_{av}$ and $\mathbf{B}_{dx}$ as 
\begin{align}
    \mathbf{B}_{av} = \vert \hat{\mathbf{F}} \vert 
    &&
    \mathbf{B}_{dx} = \hat{\mathbf{F}} - \mathrm{diag}(\textstyle\sum_j\hat{\mathbf{F}}_{:,j}),
\end{align}
where $\hat{\mathbf{F}}$ denotes the vector field $\mathbf{F}$ after normalising each row by its $L_1$-norm and $|\cdot |$ denotes the absolute value.
Message passing based on these aggregation schemes enables the messages to contain a notion of directionality --- an indication of where a message comes from.
$\mathbf{B}_{av}$ can intuitively be understood to smooth the signal based on the directions specified by $\mathbf{F}$. 
In contrast, the aggregation matrix $\mathbf{B}_{dx}$ computes the discrete derivative of the signal. 
The visualisation of both $\mathbf{B}_{av}$ and $\mathbf{B}_{dx}$ for a specific node in the graph is shown in Figure \ref{fig:anisotropic filters diagram}. We use $\mathbf{B}_{av}$ and $\mathbf{B}_{dx}$ to define the anisotropic filters in our architecture.
\subsection{Graph Anisotropic Diffusion Architecture}
\label{Graph Anisotropic Diffusion Architecture}
\begin{figure}
\centering
\begin{tikzpicture}[node distance=0cm]


[1.4031],
        [2.8343],
        [2.1150],
        [2.1169],
        [2.8373],
        [1.4050],
        [1.4050],
        [2.8303],
        [2.1177],
        [2.1133],
        [2.1132],
        [2.1133],
        [2.1177]

output tensor([ -8.6747,   2.7701,  -3.6068,  -4.1010,   2.2107,  -7.0348,  -5.7254,
          3.3399,   0.1202,  -7.5780,  -4.1146,  -1.3538, -11.6763]

5.7121e-02
3.2715e-01
6.1597e-02
1.7089e-02
5.4776e-03
1.2834e-03

3.6112e-01
1.0962e-01
7.4019e-02
1.0962e-01

\node (graph_node_1) [graph, fill=black!100] {};
\node (graph_node_2) [graph, shift=({-30:0.5cm}), fill=green!48] {};
\node (graph_node_3) [graph, right of=graph_node_2, shift=({30:0.5 cm}), fill=green!25] {};
\node (graph_node_4) [graph, right of=graph_node_3, shift=({90:0.5cm}), fill=green!11] {};
\node (graph_node_5) [graph, right of=graph_node_4, shift=({150:0.5cm}), fill=green!25] {};
\node (graph_node_6) [graph, shift=({90:0.5cm}), fill=green!48] {};
\draw [thick,-,-=stealth] (graph_node_2) -- (graph_node_1);
\draw [thick,-,-=stealth] (graph_node_6) -- (graph_node_1);
\draw [thick,-,-=stealth] (graph_node_3) -- (graph_node_2);
\draw [thick,-,-=stealth] (graph_node_6) -- (graph_node_5);
\draw [thick,-,-=stealth] (graph_node_3) -- (graph_node_4);
\draw [thick,-,-=stealth] (graph_node_4) -- (graph_node_5);

\node (graph_node_7) [graph, shift=({210:0.5cm}), fill=green!38] {};
\node (graph_node_8) [graph, right of=graph_node_7, shift=({-90:0.5cm}), fill=green!7] {};
\node (graph_node_9) [graph, right of=graph_node_7, shift=({150:0.5cm}), fill=green!9] {};

\draw [thick,-,-=stealth] (graph_node_7) -- (graph_node_1);
\draw [thick,-,-=stealth] (graph_node_8) -- (graph_node_7);
\draw [thick,-,-=stealth] (graph_node_9) -- (graph_node_7);

\node (graph_node_10) [graph, right of=graph_node_9, shift=({210:0.5cm}), fill=green!4] {};
\draw [thick,-,-=stealth] (graph_node_10) -- (graph_node_9);

\node (graph_node_11) [graph, right of=graph_node_10, shift=({150:0.5cm}), fill=green!2] {};
\draw [thick,-,-=stealth] (graph_node_10) -- (graph_node_11);

\node (graph_node_12) [graph, right of=graph_node_11, shift=({90:0.5cm}), fill=green!1] {};
\node (graph_node_13) [graph, right of=graph_node_11, shift=({210:0.5cm}), fill=green!1] {};

\draw [thick,-,-=stealth] (graph_node_12) -- (graph_node_11);
\draw [thick,-,-=stealth] (graph_node_13) -- (graph_node_11);

\node (value) [right of=graph_node_6, shift=({180:0.75cm})] {$av_{1}$};

\end{tikzpicture}
\hspace{0.5cm}
\begin{tikzpicture}[node distance=0cm]

second layer: output tensor([-8.0991, -2.7187, -3.4818, -4.4134,  0.7891, -4.7144, -0.5021, -2.1727, -0.5071, -0.0608, -0.0597, -1.5053,  1.1611]

5.7587e-02 -> 7
3.2982e-01 -> 38
6.2099e-02 -> 9
1.7228e-02 -> 5
5.5223e-03 -> 3
9.6421e-04 - > 1
9.6421e-04

-2.6520e-01 -> -30
-8.0502e-02 -> -13
-4.2168e-02 -> -6
-8.0502e-02

\node (graph_node_1) [graph, fill=black!100] {};
\node (graph_node_2) [graph, shift=({-30:0.5cm}), fill=red!30] {};
\node (graph_node_3) [graph, right of=graph_node_2, shift=({30:0.5 cm}), fill=red!13] {};
\node (graph_node_4) [graph, right of=graph_node_3, shift=({90:0.5cm}), fill=red!6.35] {};
\node (graph_node_5) [graph, right of=graph_node_4, shift=({150:0.5cm}), fill=red!13] {};
\node (graph_node_6) [graph, shift=({90:0.5cm}), fill=red!30] {};
\draw [thick,-,-=stealth] (graph_node_1) -- (graph_node_2);
\draw [thick,-,-=stealth] (graph_node_1) -- (graph_node_6);
\draw [thick,-,-=stealth] (graph_node_2) -- (graph_node_3);
\draw [thick,-,-=stealth] (graph_node_5) -- (graph_node_6);
\draw [thick,-,-=stealth] (graph_node_5) -- (graph_node_4);
\draw [thick,-,-=stealth] (graph_node_3) -- (graph_node_4);

\node (graph_node_7) [graph, shift=({210:0.5cm}), fill=green!38] {};
\node (graph_node_8) [graph, right of=graph_node_7, shift=({-90:0.5cm}), fill=green!7] {};
\node (graph_node_9) [graph, right of=graph_node_7, shift=({150:0.5cm}), fill=green!9]  {};

\draw [thick,-,-=stealth] (graph_node_7) -- (graph_node_1);
\draw [thick,-,-=stealth] (graph_node_8) -- (graph_node_7);
\draw [thick,-,-=stealth] (graph_node_9) -- (graph_node_7);

\node (graph_node_10) [graph, right of=graph_node_9, shift=({210:0.5cm}), fill=green!4] {};
\draw [thick,-,-=stealth] (graph_node_10) -- (graph_node_9);

\node (graph_node_11) [graph, right of=graph_node_10, shift=({150:0.5cm}), fill=green!2] {};
\draw [thick,-,-=stealth] (graph_node_10) -- (graph_node_11);

\node (graph_node_12) [graph, right of=graph_node_11, shift=({90:0.5cm}), fill=green!1] {};
\node (graph_node_13) [graph, right of=graph_node_11, shift=({210:0.5cm}), fill=green!1] {};

\draw [thick,-,-=stealth] (graph_node_12) -- (graph_node_11);
\draw [thick,-,-=stealth] (graph_node_13) -- (graph_node_11);

\node (value) [right of=graph_node_6, shift=({180:0.75cm})] {$dx_{1}$};

\end{tikzpicture}
\hspace{0.5cm}
\begin{tikzpicture}
\pgfplotscolorbardrawstandalone[ 
    colormap name = my colormap,
    colorbar horizontal,
    point meta min=-1,
    point meta max=1,
    colorbar style={
        width=2cm,
        height=0.2cm,
        rotate=90,
        x tick label style={font=\tiny, xshift = 0.1cm, yshift = 0.2cm, text width=0.05cm,align=right},
        xtick={0},
        }]
        
\node (max) [xshift= 0.85cm, yshift = -0.1cm] {\small{max}};
\node (min) [xshift= 0.85cm, yshift = -1.9cm] {\small{min}};
\end{tikzpicture}

    \caption{Anisotropic diffusion kernels for the black node. Diffusion and anisotropic filter have been applied to a one-hot vector in each node and observed the change in the black node. (The time $t$ is the largest channel time in the first layer of GAD architecture trained on ZINC).}
    \label{fig:Global Anisotropic Diffusion kernel}
\end{figure}
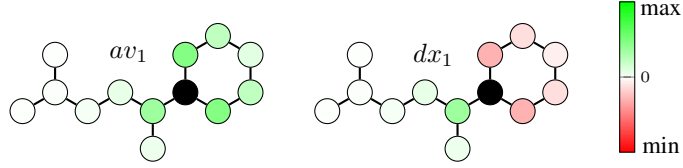
We combine the diffusion layer in Section \ref{Learned Diffusion} and the aggregation matrices in Section \ref{Anisotropic filters} in one architecture which we call Graph Anisotropic Diffusion (GAD). One block of GAD is shown in Figure \ref{One block of Graph Anisotropic Diffusion}.
The input node features are passed through a diffusion layer $h_{t}$. 
Next, we apply a non-linearity to the output of the diffusion and then the aggregation matrices ($\mathbf{B}_{av}$ and $\mathbf{B}_{dx}$).
Last, the output of the diffusion layer is concatenated with the output of these aggregation matrices and passed through an MLP.
We also use a skip connection in every GAD block, as shown in Figure \ref{One block of Graph Anisotropic Diffusion}. By combining the linear diffusion and the local anisotropic filters, we can obtain a global anisotropic diffusion, as shown in Figure \ref{fig:Global Anisotropic Diffusion kernel}.
\section{Related work}
\label{Related work}
\textbf{Image processing.} 
\label{Image processing, computer vision, and graphics}
Diffusion equations have been widely used for image enhancement, filtering, and denoising. The heat kernel of the isotropic diffusion equation on a Euclidean domain is a Gaussian, acting as a low-pass filter. More advanced non-homogeneous \citep{perona1990scale}, anisotropic \citep{weickert1998anisotropic}, or non-Euclidean \citep{661181} diffusion PDEs were studied in the context of variational and PDE-based image processing techniques.  
%
The popular bilateral filtering technique of  \citet{710815} and \citet{10.1145/566654.566574} is directly related to the latter.  

\textbf{Computer graphics and geometry processing.}
\label{Anisotropic diffusion on surfaces}
In computer graphics, the heat kernel of the diffusion equation on a surface is related to its Gaussian curvature. This property was exploited by  \citet{10.5555/1735603.1735621,5539838} to construct heat kernel signature (HKS) shape descriptors. 
\citet{andreux2014anisotropic} introduced the use of the anisotropic Laplace-Beltrami operator to give a directional diffusion operation on discrete meshes. 
\citet{boscaini2016anisotropic} also proposed a convolutional neural network architecture based on anisotropic diffusion kernels to learn an intrinsic dense correspondence between non-rigid shapes.
More recently, \citet{sharp2021diffusionnet} combined directional filters with diffusion for learning on meshes and point clouds.

\textbf{Differential equations on graphs.}
\label{Neural differential equations on graphs}
Diffusion processes and random walks on graphs have long been exploited, including the classical PageRank algorithm \citep{page1999pagerank} (steady state of diffusion on the web graph) that powered the early version of Google search engine ranking, or non-linear dimensionality reduction by diffusion maps proposed by \cite{coifman2006diffusion}.   
%
%
%
In the graph ML literature, GNNs were formulated as ordinary differential equations (ODEs) by \citet{avelar2019discrete,poli2021graph},
continuous diffusion-type processes  \citep{xhonneux2020continuous, gu2021implicit},
or physically-informed architectures  
\citep{sanchezgonzalez2019hamiltonian}.
On the other hand, solving the non-Euclidean diffusion equations using the Laplacian eigenvectors and eigenvalues was considered a crucial tool for some early techniques on graph neural networks \citep{henaff2015deep, kipf2017semisupervised, levie2018cayleynets}.
GRAND \citep{chamberlain2021grand} formulated attention-type GNNs as discretisations of the diffusion equation. BLEND \citep{chamberlain2021beltrami} proposes a non-Euclidean extension to GRAND. The key difference between our work and GRAND or BLEND \citep{chamberlain2021grand, chamberlain2021beltrami} is that we use a closed-form solution to graph diffusion, interleaved with an anisotropic filter, rather than a numerical solver. 

\textbf{Solving PDEs using neural networks.}
\label{Neural networks for solving PDEs}
\citet{raissi2017physics} used neural networks for the solution of non-linear PDEs. Efficient PDE solvers were implemented using Fourier neural operator  \citep{li2021fourier}. A similar idea has been implemented using GNNs \citep{li2020multipole}.
\citet{belbuteperes2020combining} used a GNN for predicting fluid flow and incorporated the PDE within the network. Our work is different from these techniques since they solve a given PDE using neural networks, whereas we use a PDE as a `physical metaphor' and a way to represent the space of solutions produced by a GNN.  
\section{Experiments and results}
\label{Experiments and Results}
To evaluate  our model, we tested it on two popular benchmarks: ZINC \citep{ZINC} and coordinate-free QM9 \citep{qm9}. The implementation using the PyTorch \citep{paszke2019pytorch} and PyTorch Geometric \citep{fey2019fast} libraries is publicly available at \href{https://github.com/Ahmed-A-A-Elhag/GAD}{https://github.com/Ahmed-A-A-Elhag/GAD}. Our results on each dataset are presented with a comparison with previous models in Sections \ref{ZINC} and \ref{QM9}. Appendix \ref{Hyperparameter Setting} contains the hyperparameter settings.

Following \cite{beaini2021directional}, in our implementation of the local filters, we use mean, max, and min aggregators on top of $\mathbf{B}_{av}$ and $\mathbf{B}_{dx}$, followed by the degree scalers of PNA \citep{corso2020principal}. 

\subsection{ZINC}
\label{ZINC}
ZINC is a molecular graph dataset in which the task is to predict the penalised water-octanol partition coefficient of a certain molecule (formulated as a graph regression task). We use the splits and evaluation criteria provided by \cite{dwivedi2020benchmarking}: 10K training, 1K validation, and 1K test, with a maximum of 100K parameters and mean absolute error (MAE) as evaluation metric. We compare our model against various GNNs  architectures: GCN \citep{kipf2017semisupervised},
GIN \citep{xu2019powerful},
GraphSage \citep{hamilton2018inductive},
GAT \citep{GAT},
MoNet \citep{monti2016geometric},
GatedGCN \citep{bresson2018residual},
PNA \citep{corso2020principal},
DGN \citep{beaini2021directional},
HIMP \citep{fey2020hierarchical},
SMP \citep{vignac2020building},
and GSN \citep{bouritsas2021improving}.
We ran each solving scheme (implicit timestep and spectral expansion) five times to estimate the  mean performance and its standard deviation. Our results are provided in Table \ref{Mean absolute error (MAE): ZINC dataset}. 
\begin{table}[ht]
\caption{Mean absolute error (MAE) for ZINC. GAD-i uses the implicit timestep approach and GAD-s the spectral expansion scheme. *Uses additional tailored features such as cycles.}
\label{Mean absolute error (MAE): ZINC dataset}
\begin{center}
\begin{small}
\begin{sc}
\begin{tabular}{lcccr}
\toprule
            Model &  No edge features &  Edge features\\
            
            \midrule 
            GCN    & 0.469$\pm$ 0.002&  -\\
            GIN & 0.408$\pm$ 0.008& - \\
            GraphSage    & 0.410$\pm$ 0.005& -\\
            GAT    & 0.463$\pm$ 0.002& -      \\
            MoNet     & 0.407$\pm$ 0.007& -\\
            GatedGCN      & 0.422$\pm$ 0.006 & 0.363 $\pm$ 0.009 \\
            PNA      & 0.320$\pm$ 0.032& 0.188$\pm$ 0.004 \\
            \textbf{DGN}  & 0.219$\pm$ 0.010& 0.168$\pm$ 0.003\\
            HIMP*   & - & 0.151$\pm$ 0.006 \\
            SMP*   & 0.219$\pm$ & 0.138$\pm$ \\
            GSN*   & 0.140$\pm$ 0.006& 0.115$\pm$ 0.012\\
            \midrule 
            \textbf{GAD-i}  & 0.181$\pm$ 0.004& 	0.139$\pm$ 0.007\\
            \textbf{GAD-s}   & 0.194$\pm$ 0.006& 0.140$\pm$ 0.004\\
\bottomrule
\end{tabular}
\end{sc}
\end{small}
\end{center}
\end{table}
\subsection{QM9}
\label{QM9}
QM9 is a common molecular graph benchmark, which provides different quantum mechanical properties of small molecules. we test our model on the first seven properties provided by the PyTorch Geometric \citep{fey2019fast} library keeping the number of parameters constant and removing the 3D coordinates from the model's inputs. For the data split, we use around 110k training, 10k validation, and 10k test samples as provided by \cite{pmlr-v119-brockschmidt20a}. The evaluation metric is mean absolute error (MAE). Our results on QM9's properties are provided in Table \ref{Mean absolute error (MAE): QM9 properties}. 
\begin{table}[ht]
\caption{Mean absolute error (MAE) for QM9's properties. GAD-s uses the spectral expansion scheme.}
\label{Mean absolute error (MAE): QM9 properties}
\begin{center}
\begin{small}
\begin{sc}
\begin{tabular}{lcccccccr}
            \midrule 
             & $\mu$  & $\alpha$ & $\epsilon_{\textrm{HOMO}}$ & $\epsilon_{\textrm{LUMO}}$ & $\Delta\epsilon$ & $\langle R^2 \rangle$ & $\textrm{ZPVE}$ \\

             Unit & \textrm{D} & $a_{0}^{3}$ & \textrm{meV} & \textrm{meV} & \textrm{meV} & ${a_0}^2$ & \textrm{meV}\\
             \midrule 
             PNA            &   0.365 & 0.255 & 71.5 & 70.1 & 100.2 & 18.3 & 6.57 \\
             \textbf{DGN}   &  0.354 & 0.250 & 71.1 & 68   & 99.2  & 17.2 & 6.03 \\ 
             \textbf{GAD-s} &  0.338 & 0.240 & 64.3 & 63   & 96.8  & 16.5 & 4.84 \\
\bottomrule
\end{tabular}
\end{sc}
\end{small}
\end{center}
\end{table}

Given that DGN \citep{beaini2021directional} is the baseline GNN that our model was constructed upon, we consider this as the main comparison point to analyse the effectiveness of the diffusion process presented. On ZINC, the addition of the diffusion layer reduces the MAE by more than 17\%, and on QM9, it improves the performance for every single property. Note that while methods using the addition of tailored features or message passing structure achieve better performances on ZINC, these improvements are orthogonal to the ones proposed in this work and could be combined.
\section{Conclusion}
\label{Conclusion}
We proposed Graph Anisotropic Diffusion (GAD), a novel method that consists of a new efficient way of performing isotropic diffusion on graphs combined with local anisotropic kernels. This approach is used to obtain a global anisotropic diffusion. Empirical results on popular molecular property prediction benchmarks confirm the improved predictive capacity of our method. We hope this work spurs future research on alternative methods to obtain anisotropic diffusion on graphs and theoretical analyses of their expressive power.  
\clearpage






\bibliography{iclr2022_workshop}
\bibliographystyle{iclr2022_workshop}

\newpage

\appendix


\section{Hyperparameter Settings}
\label{Hyperparameter Setting}
For ZINC, the total number of parameters used in GAD is the same as the one used in DGN \citep{beaini2021directional} ($\approx$ 100K). We also use the same hyperparameter settings in DGN \citep{beaini2021directional} and decrease the hidden dimensions to keep the same number of parameters. In the spectral expansion scheme, we use the number of eigenvectors $ k \in \{ 20, 25, 30 \} $.

For QM9, we fine-tuned GAD over the following hyperparameter settings:

\begin{itemize}
    \item Weight decay $ \in \{ 3 \times 10^{-3}, 3 \times 10^{-4}, 
    3 \times 10^{-5}, 3 \times 10^{-6}, 3 \times 10^{-7} \} $.
    \item Dropout $ \in \{ 0, 0.1, 0.2, 0.3 \} $.
    \item $ lr \in \{ 1 \times 10^{-3} \} $.
    \item Batch size $ \in \{ 128, 256 \} $.
    \item Number of layers $ \in \{ 6, 7, 8, 9, 10 \} $.
    \item Hidden dimensions $ \in \{ 75, 90, 130, 150 \} $.
    \item Number of eigenvectors $ k \in \{ 15, 20, 25 \} $.
    \item Aggregators $ \in \{ (mean, av_{1}, dx_{1}), (mean, max, min, dx_{1}), (mean, max, min, av_{1}, dx_{1}), \\ (mean, sum, max, dx_{1}) \} $.

\end{itemize}
In both the PNA \citep{corso2020principal} and DGN \citep{beaini2021directional} models, we ran them with the hyperparameters we obtained from the above search and increased the size of hidden dimensions to have the same number of parameters used in GAD.
Aggregators used in PNA \citep{corso2020principal} are $(mean, max, min, std)$, and in DGN \citep{beaini2021directional} are $(mean, sum, max, dx_{1})$.
\end{document}